\RequirePackage{etex}
\documentclass[runningheads]{llncs}

\usepackage[T1]{fontenc}
\usepackage{graphicx}
\usepackage{booktabs}
\usepackage[misc]{ifsym}
\newcommand{\corr}{(\Letter)}

\usepackage{listings}
\usepackage{url}
\urldef{\publishedDOI}\url{https://doi.org/10.1007/978-3-032-16708-8_4}
\usepackage{multirow}
\usepackage{lmodern}
\usepackage[table]{xcolor}
\usepackage{array}
\usepackage{amsmath}
\usepackage{amssymb}
\usepackage{tikz}
\usetikzlibrary{positioning}
\usepackage{tikz-network}
\usepackage{fontawesome}
\usepackage{adjustbox}
\usepackage{xspace}
\usepackage{paralist}
\usepackage{import} \usepackage{bibentry} \usepackage{pgfmath} \usepackage{siunitx}
\usepackage{soul}
\usetikzlibrary{decorations.pathreplacing} \usetikzlibrary {shapes.arrows, shadings, arrows.meta} \usetikzlibrary{shapes.geometric} 
\usetikzlibrary{shapes.symbols} 
\begin{document}


\title{RAG-Enhanced LLMs for Interactive Explainability
in Clinical Decision Support Systems\thanks{%
This is the preprint version of the conference paper published by Springer.
Please cite the published version:
Lorenzo Mannocci, Francesca Naretto, Lucia Passaro, and Anna Monreale (2026).
RAG-Enhanced LLMs for Interactive Explainability in Clinical Decision Support Systems.
\textit{Artificial Intelligence for Healthcare, and Hybrid Models for Coupling Deductive and Inductive Reasoning}.
HC\_AIxIA\_HYDRA 2025.
Communications in Computer and Information Science,
vol.~2830, pp.~50--63. Springer, Cham.
\publishedDOI}}

\titlerunning{RAG-Enhanced LLMs for Interactive Explainability in CDSS}

\author{Lorenzo Mannocci\inst{1} \corr \orcidID{0000-0002-5556-3746} \and
Francesca Naretto\inst{2}\orcidID{0000-0003-1301-7787} \and
Lucia Passaro\inst{2}\orcidID{0000-0003-4934-5344} \and
Anna Monreale\inst{2}\orcidID{0000-0001-8541-0284}
}

\authorrunning{L. Mannocci et al.}

\institute{University of Pisa, Italy \email{\{lorenzo.mannocci\}@di.unipi.it}
\and
University of Pisa, Italy \email{\{francesca.naretto, lucia.passaro, anna.monreale\}@unipi.it}
}

\maketitle              

\begin{abstract}
Artificial Intelligence (AI) is increasingly embedded in high-risks decision-making systems across domains such as healthcare, finance, and the judicial system. As reliance on predictive models grows, so does the demand for Explainable AI (XAI) to ensure transparency, trust, and interpretability. However, existing XAI approaches often fail to present explanations in a format accessible to domain experts, limiting their practical utility. Large Language Models (LLMs) have recently emerged as a bridge between complex models and end-users, offering natural language explanations. Yet, their use introduces new risks, including hallucinated outputs and lack of source traceability. To address these challenges, we propose a Clinical Decision Support System (CDSS) that not only combines standard predictive models with XAI techniques but also incorporates a Retrieval-Augmented Generation module as an interactive enhancer. This module grounds explanations in verified medical knowledge, reformulates technical outputs into a human-like format, and enables users—such as clinicians—to actively query the system. By fostering an interactive, human-in-the-loop environment, our approach empowers domain experts to explore model decisions, contextualize explanations, and build trust in AI-assisted diagnostics. The study concludes with preliminary experiments validating the proposed methodology.

\end{abstract}

\section{Introduction}
\label{sec:introduction}
Artificial Intelligence (AI) has rapidly evolved to become a crucial component in decision-making systems across various fields, including healthcare~\cite{shaheen2021applications}, finance~\cite{cao2022ai}, law enforcement~\cite{raaijmakers2019artificial}, and the judicial system~\cite{nowotko2021ai}. The increasing reliance on predictive models in these domains necessitates the development of methods to explain their outputs, a requirement that has led to the emergence of Explainable Artificial Intelligence (XAI)~\cite{guidotti2018survey,giannotti2023explainable}. XAI aims to make AI models more interpretable, transparent, and trustworthy, ensuring that domain experts and end-users can understand and assess the rationale behind automated decisions~\cite{doshi2017towards,miller2019explanation}. The need for trust and transparency in AI decision-making has also been recognized at the regulatory level. The European Union’s AI Act~\cite{eu2024ai}, one of the first comprehensive legal frameworks for AI, emphasizes the importance of explainability, particularly for high-risk AI applications. The AI Act mandates that AI systems used in critical sectors provide clear, understandable, and verifiable explanations to ensure accountability and fairness. This regulatory push further underscores the necessity of effective XAI techniques that can meet both ethical and legal standards while supporting domain experts in making informed decisions.

Many high-risk applications depend on predictive models that significantly impact human lives. In healthcare, AI-driven diagnostic systems assist doctors in detecting diseases, recommending treatments, and predicting patient outcomes~\cite{shaheen2021applications}. In the judicial system, predictive models inform parole decisions and risk assessments~\cite{nowotko2021ai}. Despite their effectiveness, these models often operate as black boxes, making it difficult for domain experts to verify their outputs and detect potential biases. Without proper interpretability, decision-makers may struggle to trust AI-driven recommendations, leading to ethical, legal, and social concerns. Indeed, these explanations might be difficult to interpret, particularly for professionals outside the field of computer science, such as doctors, lawyers, or policymakers. The challenge lies in presenting these explanations in a way that aligns with the expertise and expectations of the end-users~\cite{conati2021toward}. For example, rule-based explainability methods may generate decision rules with numerical thresholds that do not align with conventional domain knowledge. In medical applications, an AI model might define a risk threshold for blood pressure or glucose levels that deviates from commonly accepted medical guidelines, creating confusion among clinicians. Without proper contextualization, such outputs can diminish trust in the AI system rather than enhancing its credibility.
Moreover, existing XAI methods often fail to present explanations in a format that is intuitive and easily digestible for domain experts. This limitation has led to the integration of Large Language Models (LLMs) as interfaces for explainability, offering a more natural and human-like explanations for AI predictions~\cite{bilal2025llms}.

LLMs have been increasingly employed as mediators between complex AI models and domain experts, translating complex mathematical justifications into human-like explanations. By reformatting and simplifying the outputs of XAI techniques, LLMs can substantially enhance the interpretability of AI-driven decisions~\cite{bilal2025llms}. In a healthcare setting, for instance, an LLM can transform a technical explanation of a machine learning model's prediction into a structured summary that aligns with a physician's diagnostic workflow~\cite{ichmoukhamedov2024good,mavrepis2024xai}. More recently, LLMs have been leveraged not only to translate explanations into natural language but also to enrich them with contextual information drawn from their pre-trained knowledge~\cite{park2025synergistic}. However, this approach introduces an additional layer of complexity and potential risk. LLMs are known to occasionally produce inaccurate or misleading outputs, a phenomenon referred to as hallucinations~\cite{ji2023survey}. Furthermore, these models typically do not provide access to the underlying sources that support their responses, limiting transparency and traceability~\cite{li2023survey}.

One promising approach to mitigating LLM hallucinations in XAI is the use of Retrieval-Augmented Generation (RAG)~\cite{lewis2020retrieval}. RAG enhances LLM-generated explanations by grounding them in an external, domain-specific knowledge base. Instead of relying solely on the LLM’s pre-trained knowledge, RAG retrieves relevant information from a curated database to provide more accurate and reliable explanations. In the context of XAI, RAG can address two critical challenges. First, it reduces the risk of hallucinations by ensuring that explanations are based on verified sources rather than probabilistic text generation. Second, it allows for the integration of domain-specific knowledge, ensuring that explanations adhere to established medical, legal, or financial guidelines. By leveraging RAG, AI-generated explanations can be aligned more closely with experts' expectations, thereby increasing trust and usability in real-world applications.

In this work, we present, to the best of our knowledge, the first Clinical Decision Support System (CDSS) that integrates standard predictive models with XAI techniques and RAG-enhanced explanations. The proposed system consists of three key components:

\begin{enumerate}
    \item \textit{Predictive Model}: A standard classifier, such as a neural network or a random forest, that generates predictions based on patient data;
    \item \textit{Explainer}: An explanation method that provides insights into the model’s decision-making process through feature importance-based explanations (e.g., LIME~\cite{ribeiro2016should} and SHAP~\cite{lundberg2017unified}), decision rules (e.g., LORE~\cite{GuidottiMRNTPG24}) or exemplars and counter-exemplars ~\cite{Guidotti24};
    \item \textit{RAG-Based enhancer module}: 
    A module built on RAG that enhances explanations by grounding them in reliable medical knowledge, mitigating hallucinations, and transforming traditional explanations into more human-readable, context-aware formats. It also serves as an interactive tool, allowing end users to ask specific questions and explore the patient’s clinical profile in relation to the model’s predictions and explanations.
\end{enumerate}

By combining standard classifiers with RAG-based explanations grounded in verified medical sources, our system enhances the trustworthiness and domain relevance of AI-driven medical decisions. This approach addresses key limitations of conventional XAI by providing interpretable, reliable insights tailored to healthcare professionals. The study concludes with \textit{preliminary experiments} validating the proposed methodology.

\section{Related Works}
\label{sec:related_works}

AI is transforming modern society, largely due to the rise of complex Machine Learning (ML) models with strong predictive capabilities. However, these models—such as deep neural networks and ensemble methods—are often considered black boxes due to their lack of transparency~\cite{miller2019explanation,guidotti2018survey}. This opacity makes it difficult to identify biases in training data or explain decisions, which is especially concerning in high-stakes domains like healthcare, finance, and criminal justice. As Rudin argues~\cite{rudin2019stop}, the lack of interpretability hinders trust, accountability, and real-world deployment of AI systems in regulated settings. At the same time, Large Language Models (LLMs) have opened new possibilities for generating intuitive, accessible explanations. Their ability to bridge technical and non-technical understanding supports the development of more human-centric, fair, and transparent AI systems.

We focus next on two key areas: post-hoc explanations, which analyze model behavior after predictions are made, and human-centered explanations, which adapt to user needs to improve interpretability.

\subsection{Post-hoc explanations}
In the context of post-hoc explanation methods for black-box models applied to tabular data, both local and global approaches have been extensively studied and benchmarked~\cite{bodria2023benchmarking}. Local explanation methods aim to interpret individual predictions by approximating the model’s behavior in the vicinity of a specific instance. One of the most widely known techniques in this category is LIME, which builds a simple surrogate model—usually linear—on a locally perturbed dataset to estimate the contribution of each feature~\cite{ribeiro2016should}. Although popular, LIME suffers from instability due to its random neighborhood generation strategy. A more theoretically grounded alternative is SHAP, which relies on Shapley values from cooperative game theory to assign each feature an importance score based on its marginal contribution across all possible feature combinations~\cite{lundberg2017unified}. SHAP ensures consistency and local accuracy, making it a robust choice for feature attribution. Another relevant method is LORE, which employs genetic algorithms to generate a neighborhood and extracts interpretable decision rules, offering both factual and counterfactual explanations~\cite{GuidottiMRNTPG24}.  Other approaches provide local explanations which are based on exemplars or better counter-exemplars, which enable a counterfactual reasoning that in some sense produce suggestions to users on what is possible to change for obtaining a different decision from the machine learning model \cite{Guidotti24}. 

On the other hand, global explanation methods focus on understanding the model’s logic over the entire input space. RuleFit combines decision rules extracted from tree ensembles with linear models to produce sparse, interpretable explanations of model behavior. Building upon this idea, SkopeRules uses ensemble techniques to extract a set of high-precision rules that summarize the model’s decision boundaries~\cite{friedman2008predictive}. More recently, Setzu et al.~\cite{setzu2021glocalx} proposed GLocalX, which aggregates local explanations to form a coherent global understanding of the model, effectively bridging the gap between local interpretability and global insight. These methods, differing in their assumptions and strategies, collectively contribute to the growing toolbox of post-hoc XAI techniques tailored for tabular data.

\subsection{Human-centered explanations}
{Both post-hoc and intrinsic XAI methods have advanced model interpretability, yet a key challenge remains: ensuring explanations are truly comprehensible to end users. Human-centered XAI addresses this by aligning explanations with user needs through participatory design methods such as interviews and user studies~\cite{amershi2019guidelines,eiband2018bringing,ehsan2020human}. This approach emphasizes contextualized explanations that integrate seamlessly into user workflows and are validated through real-world testing~\cite{jiang2024human,wolf2019explainability}.

Recent work explores how LLMs can enhance explanation quality by generating adaptive, narrative-style outputs that combine predictions with counterfactuals~\cite{martens2025tell,ichmoukhamedov2024good}. For example, x-[plAIn]~\cite{mavrepis2024xai} uses a customized LLM to generate concise, audience-specific summaries of XAI methods. Narrative explanations are also being explored across data types, including tabular data\cite{burton2023natural,zytek2024llms}, graphs~\cite{giorgi2024natural,pan2024tagexplainer}, and images~\cite{wojciechowski2024faithful}.

Initially framed as a data-to-text task, LLM-generated narratives have evolved to incorporate broader knowledge, adding context and reasoning beyond simple summaries~\cite{martens2025tell}. A recent pilot study found users preferred such narratives over SHAP plots across several metrics~\cite{zytek2024llms}. This approach aligns with the concept of \textit{self-explanations}, where models articulate reasoning to aid understanding~\cite{huang2023can}. However, it also introduces risks like adversarial helpfulness, where convincing but incorrect explanations may mislead users~\cite{ajwani2024llm}. RAG helps mitigate this by grounding outputs in trusted sources, improving factual accuracy and transparency~\cite{garigliotti2024explaining}. Similarly, Park et al.~\cite{park2025synergistic} combine LLMs with knowledge graphs to generate structured, domain-specific explanations that are both clinically accurate and contextually rich.

\section{Method}
\label{sec:method}
\def\fontElement{\normalsize}
\def\fontText{\Large}
\def\fontTitle{\LARGE}

\def\lenArrow{1}
\def\wArrow{1}
\def\yTextShift{3pt}
\def\yLines{2.3}
\def\xBase{-1.2}

\begin{figure}[!htbp]
    \centering
    \scalebox{0.7}{    
        \begin{tikzpicture}[inner sep=0pt, outer sep=0pt]

\draw[line width=\wArrow, ->, -latex] (\xBase,-1 -\lenArrow) -- (\xBase,-1) 
    node[pos=0, yshift=-0.3cm] {\fontElement $(x_i, y_i)$};

\node at (\xBase,0) {\fontsize{50}{30}\selectfont \faDatabase};

\draw[line width=\wArrow, ->, -latex] (\xBase, 1) -- (\xBase,1+\lenArrow) 
    node[midway, left, xshift=-\yTextShift] {\fontElement $x_i$};
    
\draw[line width=\wArrow, ->, -latex] 
    (\xBase, 1+0.5*\lenArrow) -- (\xBase + 2*\lenArrow, 1+0.5*\lenArrow) node[midway, above, yshift=\yTextShift] {\fontElement $x_i$};

\node at (\xBase, 2+\lenArrow) {\includegraphics[width=2cm]{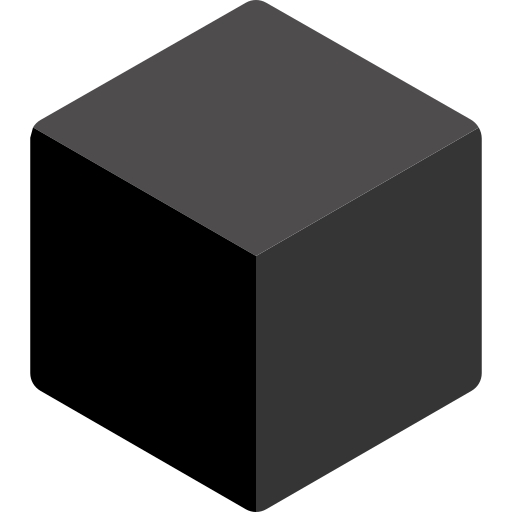}};
    \node at (\xBase,2.5+\lenArrow) [white] {\fontElement $BB$};

\draw[line width=\wArrow, ->, -latex] 
    (\xBase, 3+\lenArrow+0.1) -- (\xBase, 3+2*\lenArrow+0.1) 
    node[midway, left, xshift=-\yTextShift] {\fontElement $\tilde{y_i}$};

\draw[line width=\wArrow, ->, -latex] 
    (\xBase, 3+1.5*\lenArrow+0.1) -- (\xBase+2*\lenArrow, 3+1.5*\lenArrow+0.1) node[midway, above, yshift=\yTextShift] {\fontElement $\tilde{y_i}$};

\node at (\xBase, 4+2*\lenArrow+0.1) {\includegraphics[width=2cm]{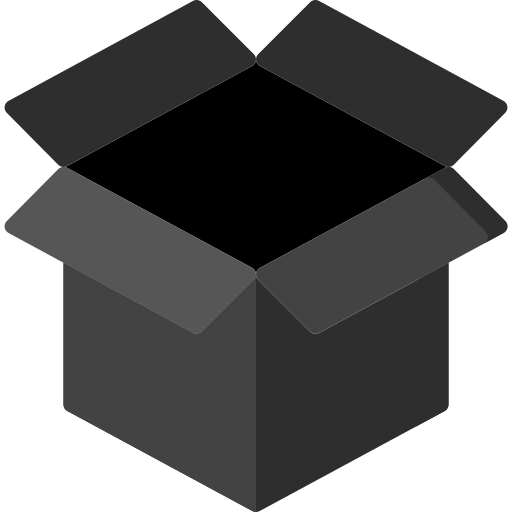}};
    \node at (\xBase,4+2*\lenArrow+0.1+ 0.35) [white] {\fontElement $XAI$};

\draw[line width=\wArrow, ->, -latex] 
    (\xBase+1, 4+2*\lenArrow+0.1) -- (\xBase+1+\lenArrow, 4+2*\lenArrow+0.1) node[midway, above, yshift=\yTextShift] {\fontElement $e_i$};

\draw[black,line width=0.1, dashed] 
    (2*\lenArrow,-1.8 -\lenArrow) 
    rectangle (5*\lenArrow, 4+3*\lenArrow+0.1);

\node[draw=black, fill=gray!10, minimum width=0.7cm, minimum height=5cm, align=center] at (\xBase+1+\lenArrow+0.35,3.8)
    {\fontElement i\\n\\p\\u\\t};

\draw[line width=\wArrow, ->, -latex] 
    (\xBase+1+\lenArrow+0.7, 4) -- (\xBase+1+\lenArrow+0.7+0.5*\lenArrow, 4);

    \node at (3.5,-\yLines) {\fontText enhancer};
    
\node at (3.5*\lenArrow, -\yLines+1.2+0.8) {\includegraphics[width=2cm]{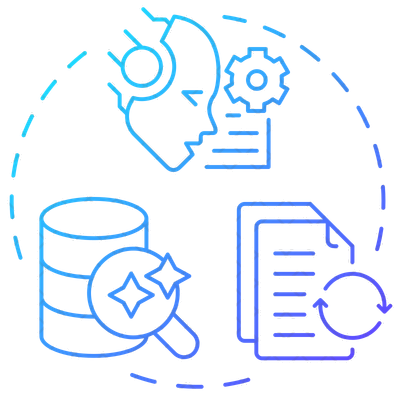}};
    \node at (3.5,-\yLines+0.8) {\fontElement RAG};

\draw[black,line width=0.1] 
    (2.5*\lenArrow-0.3,-\yLines+0.8 + 2.3) 
    rectangle (4.5*\lenArrow+0.3, -\yLines+0.45);

    \node[
        shape=cloud, 
        draw=black, 
        fill=gray!20, cloud puffs=10, 
        cloud ignores aspect, 
        minimum width=2.65cm, 
        minimum height=1.5cm, 
        align=center
    ] at (3.5*\lenArrow,6) {\fontElement question \\ \& answer};

    \node[
        shape=cloud, 
        draw=black, 
        fill=gray!20, cloud puffs=10, 
        cloud ignores aspect, 
        minimum width=2.65cm, 
        minimum height=1.5cm, 
        align=center
    ] at (3.5*\lenArrow,4) {\fontElement context \\ explanation};
    
        \node[
        shape=cloud, 
        draw=black, 
        fill=gray!20, cloud puffs=10, 
        cloud ignores aspect, 
        minimum width=2.65cm, 
        minimum height=1.5cm, 
        align=center
    ] at (3.5*\lenArrow,2) {\fontElement classifier \\\& xai};

\draw[line width=\wArrow, ->, -latex] 
    (5*\lenArrow, 2) -- (6*\lenArrow, 2) node[midway, above, yshift=\yTextShift] {\fontElement $y_i^r, e_i^r$};

\draw[line width=\wArrow, ->, -latex] 
    (5*\lenArrow, 4) -- (6*\lenArrow, 4) node[midway, above, yshift=\yTextShift] {\fontElement $e_i^e$};

\draw[black,line width=0.1, dashed] (6*\lenArrow,-1.8 -\lenArrow) rectangle (9*\lenArrow,4.5);
 
\node [align=center] at (7.5*\lenArrow,-\yLines)  {\fontText selector};

\node at (7.5*\lenArrow, -\yLines + 1) {\fontsize{30}{30}\selectfont \faCogs};
    
\node[draw, diamond, minimum size=1.9cm, align=center] at (7.5*\lenArrow, 1.5) {\fontElement is \\ $\tilde{y}=\tilde{y}^r$};

    \draw[line width=\wArrow, ->, -latex] 
    (7.5*\lenArrow, 2.5) -- 
    (7.5*\lenArrow, 2.5+0.7*\lenArrow) -- 
    (11.4*\lenArrow, 2.5+0.7*\lenArrow) 
    node[pos=0.1, below, yshift=-\yTextShift-6] {\fontElement no} 
    node[pos=0.69, above, yshift=\yTextShift] {\fontElement $(\tilde{y_i}, e_i^e)$} 
    node[pos=0.69, below, yshift=-\yTextShift, align=center] {\fontElement enhancer \\ not available};
    
    \draw[line width=\wArrow, ->, -latex] 
    (7.5*\lenArrow, -0.5+\lenArrow) -- 
    (7.5*\lenArrow, -0.5+\lenArrow-0.7*\lenArrow) -- 
    (11.4*\lenArrow, -0.5+\lenArrow-0.7*\lenArrow)
    node[pos=0.1, above, yshift=\yTextShift+6] {\fontElement yes} 
    node[pos=0.69, below, yshift=-\yTextShift] {\fontElement $(\tilde{y_i}, e_i^e, e_i^r)$};

    \draw[black,line width=0.1, dashed] (11.4*\lenArrow,2.5+0.7*\lenArrow+0.7) rectangle (14*\lenArrow, -0.5+\lenArrow-0.7*\lenArrow-0.7);
 
\node at (12.7*\lenArrow, 2) {\fontsize{30}{30}\selectfont \faDesktop};
\node [align=center] at (12.7*\lenArrow,1)  {\fontText interface};

\node at (13.2*\lenArrow, 6.1) {\fontsize{40}{30}\selectfont \faUserMd};
    
\draw[line width=\wArrow, -latex] (13.2*\lenArrow-0.3, 3.2+2.25*\lenArrow) -- (13.2*\lenArrow-0.3, 3.2+0.7*\lenArrow);

\draw[line width=\wArrow, latex-] (13.2*\lenArrow+0.3, 3.2+2.25*\lenArrow) -- (13.2*\lenArrow+0.3, 3.2+0.7*\lenArrow);

\draw[line width=\wArrow, latex-latex] 
    (5*\lenArrow, 6) -- (12*\lenArrow, 6) -- (12*\lenArrow, 3.2+0.7*\lenArrow)
    node[midway, above, xshift=-10] {\fontElement $q, a$};

\end{tikzpicture}
    }
\caption{Pipeline of the proposed Clinical Decision Support System (CDSS), integrating a predictive model, an explainer, a RAG-based enhancer that enriches and contextualizes explanations using external medical knowledge, and a selector module that governs information delivery based on confidence and consistency.}
\label{fig:method}
\end{figure}
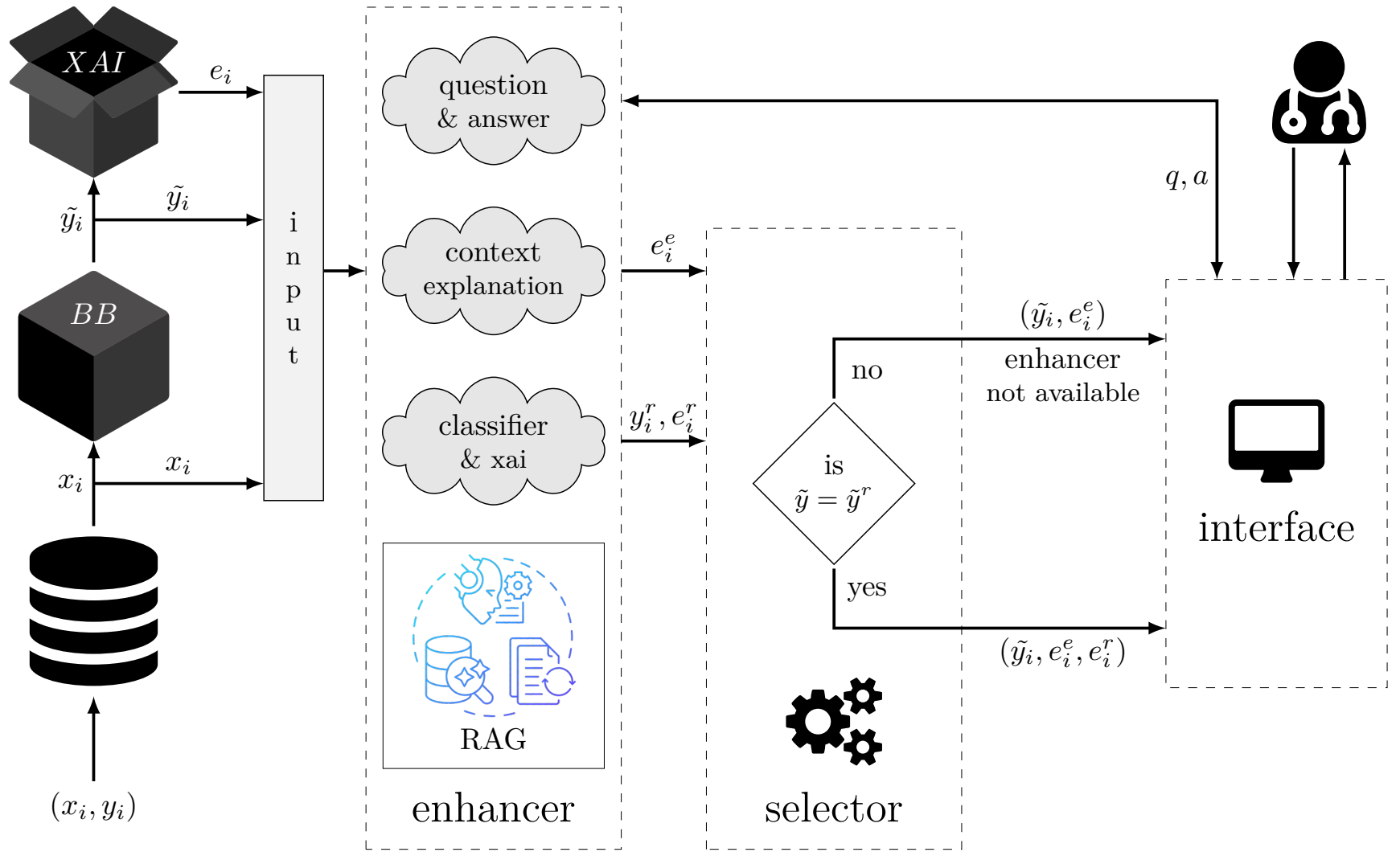 In light of the growing need for human-centered explanations and the impressive capabilities of LLMs to generate human-like responses, challenges arise, particularly when LLMs produce convincing yet incorrect answers or exhibit hallucinations. These issues can compromise the reliability of explanations and erode user trust. To address these concerns, we propose leveraging RAG, a system that combines the generative power of LLMs with the reliability of external, grounded, and reliable sources. By integrating RAG, we ensure that the explanations are not only contextually rich and human-readable but also rooted in verifiable information, thereby enhancing both accuracy and transparency in the model’s outputs. Figure~\ref{fig:method} illustrates the approach proposed in this work, which combines traditional classifiers, such as black-box models, with standard explainers like LIME or SHAP, alongside an enhancement module based on RAG. The final output of this integrated method is then processed by a selector. In the following sections, we provide a detailed description of each step in the method.

\subsection{Black-box model and explainer}
A classification task is formally defined as follows. Given a dataset $D = \{ (x_i, y_i) \}_{i=1}^n$, where $X = \{ x_1, x_2, \dots, x_n \}$ represents the input feature vectors, with $x_i \in \mathbb{R}^d$ for each $i$, and $d$ being the dimensionality of the input space, and $Y = \{ y_1, y_2, \dots, y_n \}$ is the corresponding target vector, where each $y_i \in \{ 1, 2, \dots, C \}$ represents a class label for the $i$-th data point. The goal is to learn a mapping  $f: X \to \tilde{Y},$ where $\tilde{Y}$ is the predicted class label set. The model used to perform the classification is a black-box classifier $BB$, such that the predicted output $ \tilde{y_i} = BB(x_i) $ is generated for each input $x_i$. This predicted label $\tilde{y_i}$ is then compared with the true label $y_i$ for evaluation of the model’s performance.

A local explainer is defined as a function that takes as input a specific instance $x_i$ and its predicted label $\tilde{y_i}$, and outputs an explanation $e_i$. This explanation can take various forms, such as feature importance, decision rules, exemplars, etc. Formally, a local explainer can be defined as function $\phi:(X, \tilde{Y}) \to E $, where $E$ is the explanation set. An explanation $e_i \in E$ aims to provide insight into the reasoning of the black-box classifier $BB$ for the prediction $\tilde{y_i}$ of the specific instance $x_i$.

\subsection{Enhancer}
The enhancer module, based on RAG, is designed to support three distinct tasks: \textit{context explanation}, \textit{classifier and XAI}, and \textit{question \& answer}. Each component in the pipeline operates independently and could, in principle, be implemented using LLM instead of a RAG-based approach. However, rather than relying solely on the LLM’s pre-trained knowledge, RAG retrieves relevant information from a curated medical database to generate more accurate, contextualized, and reliable explanations. \medskip

In the following we described in detail the three potential enhancer's tasks:

\paragraph{Context explanation.} 
This task enriches an existing explanation by converting the explanation into a human-like format and integrating context from the RAG’s specialized corpus. This task takes as input the feature vector $x_i$, the prediction $\tilde{y_i}$ made by the classifier, and the initial explanation $e_i$ generated by the local explainer, and provides an enhanced explanation $e_i^e$, by exploiting the RAG system. It can be formally defined as a function $C: (X, \tilde{Y}, E) \to \mathcal{E}$, where $\mathcal{E}=\{e_1^e, \ldots, e_n^e\}$ is the \textit{enhanced explanation} set. In particular the RAG system enriches the explanation $e_i$ by incorporating contextual knowledge from the corpus and transforming it into a more human-comprehensible format. The resulting output, $e_i^e$, is an enriched, human-like explanation that offers a clearer, more detailed understanding of the model’s decision-making process, grounded in both the prediction and the context of the input data;

\paragraph{Classifier and XAI.} In this task, the RAG serves as both a classifier and an explanation generator. Formally, the task can be defined as: $R:X \to (Y^r, E^r)$, where the RAG is provided with the input feature vector $x_i$ and performs a classification along with an explanation supporting the prediction. The output consists of the predicted label $y_i^r$ and an explanation $e_i^r$.

\paragraph{Question \& answer.} 
This task is the most flexible method for interacting with the system, allowing users to request interactive explanations. It is performed at the end of the pipeline, once the user has received the initial results from the entire decision support system. In this task, a question $q$ is provided as input along with the entire system's information. 
The output $a$ is the answer to the question, which may take various forms depending on the nature of $q$. This task allows the user to query the system for additional information coming from the specialized corpus of the RAG, clarification, or deeper insights into the model’s predictions and explanations, providing an interactive and dynamic way to explore and understand the decision-making process of the AI model.

\subsection{Selector}
The final component of the proposed method is the \textit{selector}, serves as a filtering and governance layer for the information presented to the user. It receives as input the black-box model's prediction $\tilde{y_i}$, the enhanced explanation $e^e_i$, the enhancer's prediction $y_i^r$, and the enhancer's explanation $e_i^r$. The primary function of the selector is to implement the decision making strategy based on the input information, ensuring that the system refrains from presenting information when confidence is insufficient or conflicting. For example, a simple but effective implementation of the selector may involve outputting the enhancer’s explanation $e_i^r$ only if the black-box classifier's prediction $\tilde{y_i}$ matches the enhancer's prediction $y_i^r$. This strategy prioritizes trust in the black-box classifier, thereby mitigating the risk of misleading or hallucinated outputs that might arise from the RAG-based enhancer. Although this represents a minimal implementation, the design space for selector strategies is broad and can be extended to incorporate more sophisticated techniques for confidence estimation, consistency checking, and risk-aware decision-making, ultimately improving the reliability and robustness of the entire AI pipeline. 

\section{Experiments and results}
\label{sec:results}

In this section, we show some preliminary results of a specific instantiation of our methodology.

\subsection{Dataset and experimental setting}
We evaluate our proposed method using the Pima Indians Diabetes Database~\cite{smith1988using}, a publicly available dataset\footnote{\url{https://www.kaggle.com/datasets/uciml/pima-indians-diabetes-database} (accessed: 19/07/2025)} originally provided by the National Institute of Diabetes and Digestive and Kidney Diseases. The goal is to diagnostically predict the presence of diabetes in a patient based on a set of clinical attributes. Prior to training, the dataset is standardized using a standard scaler, and we applied random oversampling to balance the class distribution by increasing the number of instances in the minority class, i.e., patients diagnosed with diabetes. As black-box model, we employed a Random Forest classifier trained on an 80/20 train-test split, using a grid search and 3-fold cross-validation to optimize the hyperparameters. For local explanations, we use LIME~\cite{ribeiro2016should}, which provides feature importance scores for each prediction. The enhancer module builds on the MedRAG framework\footnote{\url{https://github.com/Teddy-XiongGZ/MedRAG} (accessed: 19/07/2025)}, introduced by~\cite{xiong2024benchmarking}, leveraging the StatPearls corpus for clinical decision support and the MedCPT retriever tailored for biomedical content. The LLM used for generation is Mixtral\footnote{\url{https://huggingface.co/mistralai/Mixtral-8x7B-Instruct-v0.1} (accessed: 19/07/2025)}. Finally, the selector filters the enhancer's output, returning the RAG-based explanation $e_i^r$ only when the enhancer’s prediction $y_i^r$ matches the black-box prediction $\tilde{y_i}$, ensuring that explanations are only provided when there is consistency across models. 

The \textit{Classifier \& XAI} task receives as input the patient's \textsc{description}, represented by the feature vector $x_i$. 
The \textit{Context explanation} task extends this input by including the black-box \textsc{prediction} $\tilde{y_i}$ and the corresponding \textsc{explanation} $e_i$ provided by the explainer (LIME in this case). Lastly, the \textit{Question \& answer} task builds upon the inputs of the \textit{Context explanation} task by incorporating a user-defined \textsc{question} $q$, which MedRAG must answer based on the available information.
In Appendix~\ref{sec:app_prompt}, we provide an example of input and output of the enhancer. Figure~\ref{fig:prompt} illustrates the prompt templates employed by the enhancer module to query MedRAG, based on the specific tasks defined in Section~\ref{sec:method}. Instead, Figure~\ref{fig:explanation} shows an example instance from the PIMA dataset, for which the selector's strategy is satisfied. In the table on the top left, we display the patient features along with the classifier's prediction $\tilde{y_i}$ and the enhanced prediction $y_i^r$ generated by the RAG system. The top right section shows the feature importance values calculated using LIME, while the bottom section illustrates the output of the enhancer module based on the RAG system. It can be observed that the explanation provided by the RAG system is consistent with the one generated by LIME. Additionally, the RAG-based explanation includes references to the relevant documents in the corpus that support the explanation, thereby enabling the validation of the RAG's output.
\begin{table}[!t]
    \centering
    \small 
    \setlength{\columnsep}{10pt}
    \caption{Classification performance comparison on test set between Random Forest and MedRAG.}
    \label{tab:performance}
    \scalebox{0.8}{
        \begin{tabular}{lcrrrrr}
            \toprule
            \textbf{model} & \textbf{diabetes} & \textbf{precision} & \textbf{recall} & \textbf{f1-score} & \textbf{accuracy} & \textbf{agreement} \\
            \midrule
            \multirow{2}{*}{Random Forest} 
              & no & 0.82 & 0.79 & 0.80 & \multirow{2}{*}{0.75} &  \multirow{4}{*}{0.82} \\
              & yes & 0.64 & 0.69 & 0.66 & \\
            \cmidrule{1-6}
            \multirow{2}{*}{MedRag} 
              & no & 0.77 & 0.80 & 0.78 & \multirow{2}{*}{0.71}\\
              & yes & 0.60 & 0.56 & 0.58 & \\
            \bottomrule
        \end{tabular}
    }
\end{table} 
\subsection{Preliminary results}

\begin{figure}[!t]
    \centering
    \includegraphics[width=0.7\textwidth]{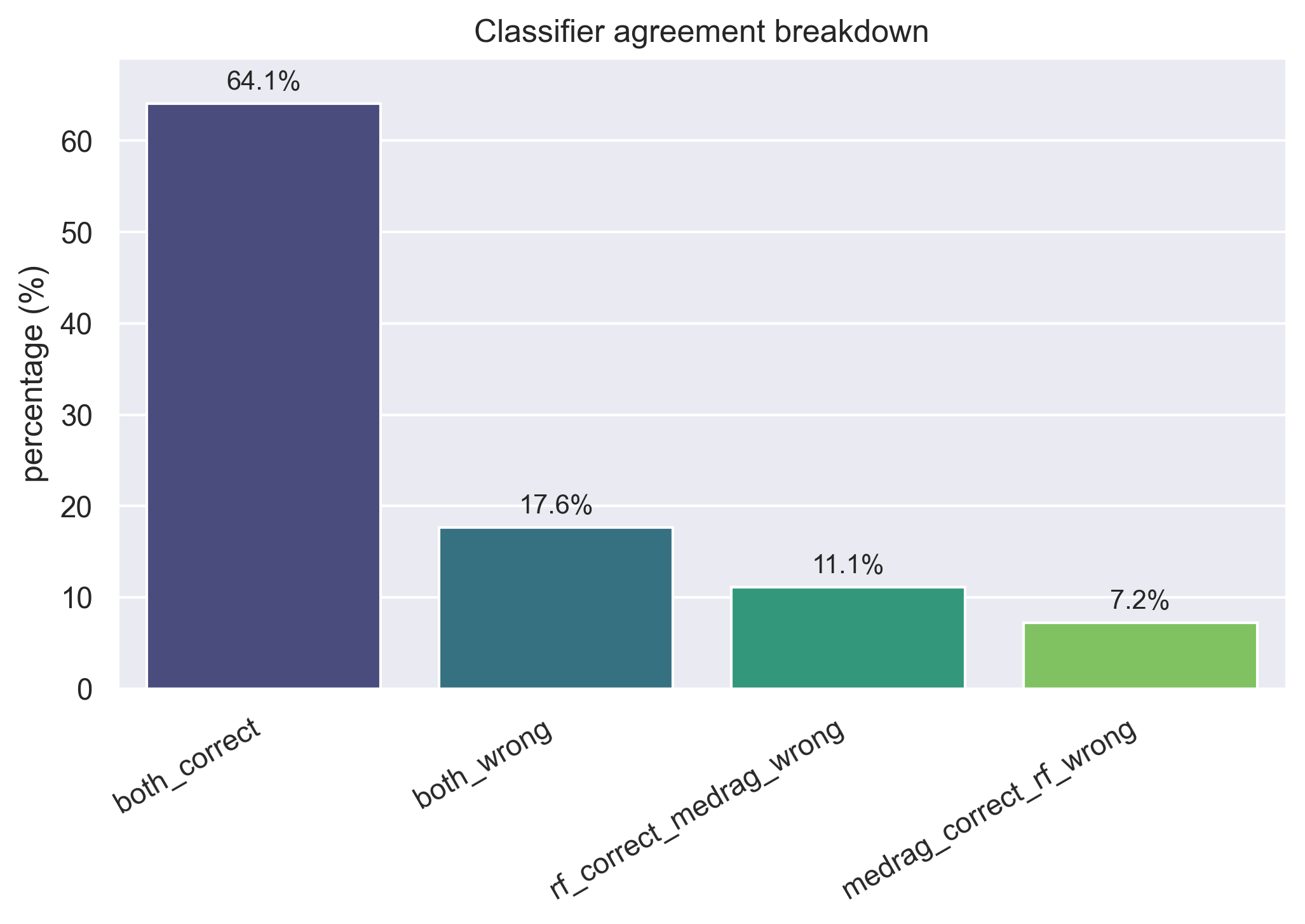}  
    \caption{Bar plot showing the agreement counts in the prediction task comparing Random Forest and MedRAG models.}
    \label{fig:agreement_breakdown}
\end{figure}

In this work, we present a preliminary evaluation of the proposed methodology to highlight the importance of integrating both a predictive model and the RAG module within the same system. To this end, we compare the predictive performance of the Random Forest (RF) classifier and the MedRAG model on the Pima Indians Diabetes dataset. The results are reported in Table~\ref{tab:performance}. Given the imbalanced nature of the dataset, with 65\% of the records indicating no diabetes, the f1-score serves as the primary evaluation metric, alongside the recall for the ``yes" class. This focus is driven by the importance of identifying as many diabetic cases as possible, even at the cost of some false positives. In this context, the Random Forest model outperforms MedRAG, particularly in the detection of diabetes cases, where it achieves an f1-score of 0.66 compared to MedRAG’s 0.58. 
Furthermore, 82\% of the time, both models agree on the prediction, which is the condition required for the selector to output the enhancer's explanation $e^r$. Figure~\ref{fig:agreement_breakdown} further breaks down this agreement, showing that the majority of instances where the models agree, they are correct. The second most frequent combination is when both models agree and are wrong. Notably, when the models disagree, the Random Forest model is more often correct than MedRAG, highlighting the greater reliability of the former in making accurate predictions.

\section{Discussion and conclusions}
Our results show that the traditional Random Forest model outperformed the RAG-based system, supporting its continued use in our CDSS. Nonetheless, the strong agreement between the two models suggests that RAG-generated explanations can complement traditional ones, enhancing interpretability and user trust.

These \textit{preliminary findings}, though promising, are limited by the small dataset, largely due to the computational cost of RAG. Expanding experiments to other clinical datasets and black-box models will help validate the system's generalizability. Future work should also explore: \textit{i)} the impact to use a LLM tailored to medical domain instead of a generalist LLM,  and \textit{ii)} how factors like model confidence and explanation alignment affect the CDSS's reliability.

Privacy remains a major concern, particularly when integrating LLMs and RAG in healthcare~\cite{naretto2020predicting,naretto2022evaluating}. Explanations often require access to sensitive clinical data, and using closed-source LLMs risks violating patient confidentiality and regulations like GDPR~\cite{yao2024survey,chadwick2024assessments}. Addressing these challenges will require privacy-preserving techniques such as local inference, federated learning~\cite{haffar2024glor,fontana2021new}, and unlearning methods~\cite{blanco2025digital}.

In summary, we propose a hybrid XAI approach that combines traditional models with generative enhancers to support interpretable, human-aligned AI in clinical contexts. While further validation is needed, this work takes a step toward more transparent and responsible decision support systems.

\begin{credits}
\subsubsection{\ackname} This research was partially supported by:
The European Commission under the NextGeneration EU programme – National Recovery and Resilience Plan (PNRR), under agreements: PNRR - M4C2 - Investimento 1.3, Partenariato Esteso PE00000013 - "FAIR - Future Artificial Intelligence Research" - Spoke 1 "Human-centered AI", and SoBigData.it – "Strengthening the Italian RI for Social Mining and Big Data Analytics” – Prot. IR0000013 – Avviso n. 3264 del 28/12/2021. The European Union’s Horizon Europe research and innovation programme for the project FINDHR (g.a. No. 101070212), and under the Excellent Science European Research Council (ERC) programme for the XAI project (g.a. No. 834756), and the HORIZON Europe projects TANGO - Grant Agreement n. 101120763
Views and opinions expressed are those of the author(s) only and do not necessarily reflect those of the European Union. Neither the European Union nor the granting authority can be held responsible for them.
\end{credits}

\bibliographystyle{splncs04}
\bibliography{mybib}
\clearpage
\appendix
\section{Input and output prompt for RAG}
\label{sec:app_prompt}
\definecolor{customBlue}{HTML}{377eb8}
\definecolor{customGreen}{HTML}{4daf4a}
\definecolor{customRed}{HTML}{e41a1c}

\def\fontElement{\normalsize}
\def\fontText{\Large}
\def\fontTitle{\LARGE}
\definecolor{customCyan}{HTML}{6ef2ff}

\def\legendY{-27}

\begin{figure}[!htbp]
    \centering
    \scalebox{0.44}{    
    \begin{tikzpicture}[inner sep=0pt, outer sep=0pt]

        \node[draw=black, fill=gray!10,text width=20cm, minimum height=10cm, align=left, inner sep=10pt] at (0,0)
        {\fontText
            \textbf{Input (classifier \& xai)} \\[1em]
            \textcolor{customRed}{You are a helpful medical expert. Your task is to classify a patient as having diabetes or not based on the provided medical data and using the relevant documents.} \\ [1em]
            
            \textcolor{customBlue}{Based on the following information: \{\textsc{description}\}, does the patient have diabetes?} \\[1em]
            
            \textcolor{customGreen}{Please provide a detailed explanation of the reasoning in ``step\_by\_step\_thinking" and a clear binary classification in ``classification" where "1" indicates ``Diabetes" and ``0" indicates "No Diabetes".
            In addition, include a section with the feature importance ranking and another section with the specific rules that influenced the classification based on each feature. In these two sections, add only relevant exploited features for the decision.
            Organize your output in JSON format as Dict``step\_by\_step\_thinking": Str(explanation), ``classification": Int(0 or 1), 
            ``feature\_importance\_ranking": List[Str], ``rules\_applied": Dict[Str, Str]. 
            Your responses will be used for research purposes only, so please provide a definite and clear answer.} \\
             };
    
            \node[draw=black, fill=gray!10,text width=20cm, minimum height=10cm, align=left, inner sep=10pt] at (0,-10.5)
            {\fontText
            \textbf{Input (context explanation)} \\ [1em]
            \textcolor{customRed}{You are an AI assistant helping interpret the output of a machine learning model that predicts whether a patient has diabetes.} \\[1em]
            
            \textcolor{customBlue}{Based on the following information: \{\textsc{description}\}. The machine learning model predicted: \{\textsc{prediction}\}. According to LIME, the most influential features for this prediction were: \{\textsc{explanation}\}.} \\ [1em]
            
            \textcolor{customGreen}{Use this information and any relevant medical background knowledge to generate a well-rounded, contextualized explanation. \\
            The explanation should describe why the model made its prediction, referencing the most influential features and their values. \\
            The explanation should be written in plain language suitable for a non-technical medical professional, and should include any relevant caveats or uncertainty if appropriate.\\}
             };
    
            \node[draw=black, fill=gray!10,text width=20cm, minimum height=10cm, align=left, inner sep=10pt] at (0,-21)
            {\fontText
            \textbf{Input (question \& answer)} \\ [1em]
            \textcolor{customRed}{You are an AI assistant helping interpret the output of a machine learning model that predicts whether a patient has diabetes. You receive a question from the doctor and you must give an answer.} \\ [1em]
            
            \textcolor{customBlue}{Based on the following information: \{\textsc{description}\}. The machine learning model predicted: \{\textsc{prediction}\}. According to LIME, the most influential features for this prediction were: \{\textsc{explanation}\}. The doctor asked the following question: \{\textsc{question}\}}. \\[1em]
            
            \textcolor{customGreen}{Use this information and any relevant medical background knowledge to provide a clear, informative answer to the doctor’s question. 
            The answer should be written in plain language suitable for a non-technical medical professional, and should include any relevant caveats or uncertainty if appropriate.\\}
             };
    
\fill[customRed] (-7,\legendY) rectangle +(1,0.2);
            \node[anchor=center, align=center, xshift=0.5cm] at (-7,\legendY-0.5) {\fontElement context/background};
            
\fill[customBlue] (0,\legendY) rectangle +(1,0.2);
            \node[anchor=center, align=center, xshift=0.5cm] at (0,\legendY-0.5) {\fontElement input information};
            
\fill[customGreen] (7,\legendY) rectangle +(1,0.2);
            \node[anchor=center, align=center, xshift=0.5cm] at (7,\legendY-0.5) {\fontElement output indicator};
    \end{tikzpicture}
    
    }
\caption{The prompt templates for the three tasks supported by the enhancer module are defined as follows. The \textit{Classifier \& XAI} task receives as input the patient's \textsc{description}, represented by the feature vector $x_i$. The \textit{Context explanation} task extends this input by including the black-box \textsc{prediction} $\tilde{y}$ and the corresponding \textsc{explanation} $e$ provided by the explainer (LIME in this case). Lastly, the \textit{Question \& answer} task builds upon the inputs of the \textit{Context explanation} task by incorporating a user-defined \textsc{question} $q$, which MedRag must answer based on the available information.}
\label{fig:prompt}
\end{figure} \definecolor{customBlue}{HTML}{377eb8}
\definecolor{customGreen}{HTML}{4daf4a}
\definecolor{customRed}{HTML}{e41a1c}
\definecolor{customViolet}{HTML}{7570b3}

\def\fontElement{\normalsize}
\def\fontText{\Large}
\def\fontTitle{\LARGE}

\def\legendY{-8}

\begin{figure}[!htbp]
    \centering
    \scalebox{0.5}{    
    \begin{tikzpicture}[inner sep=0pt, outer sep=0pt]

    \node at (-10, 16) {\fontTitle patient's record};
    
    \node[anchor=south] at (-10, 9.5) {
    \resizebox{0.5\textwidth}{!}{
        \begin{tabular}{ll}
            \toprule
            \textbf{feature} & \textbf{value} \\
            \midrule
            BloodPressure & 70.0 \\
            Pregnancies & 3 \\
            Insulin & 200 \\
            DiabetesPedigreeFunction & 0.408 \\
            SkinThickness & 22.0 \\
            Age & 36 \\
            BMI & 36.4 \\
            Glucose & 187.0 \\
            \midrule
            ground truth $y_i$ & 1 (diabetes) \\
            classifier's prediction $\tilde{y_i}$ & 1 (diabetes)\\
            enhancer's prediction $y_i^r$ & 1 (diabetes)\\
            \bottomrule
        \end{tabular}
    }
    };
        
        \node at (5, 16) {\fontTitle LIME explanation};
    
        \node at (2,12.5) {\includegraphics[width=1.5\textwidth]{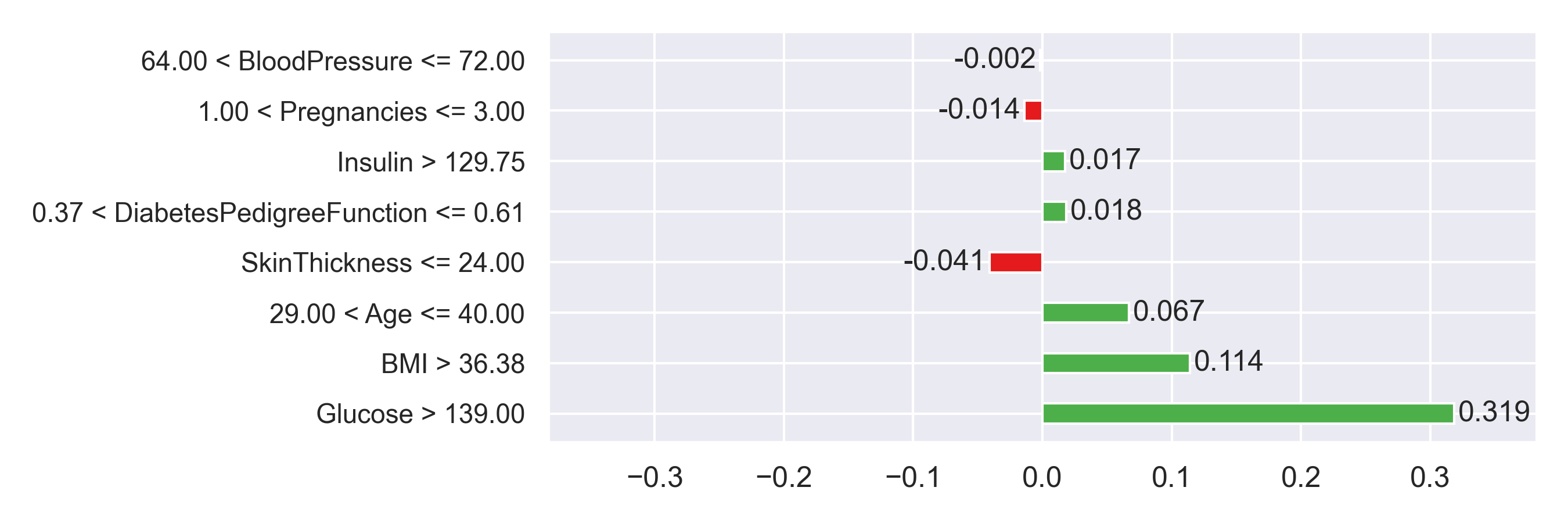}};
        
        \node[draw=black, fill=gray!10,text width=20cm, minimum height=10cm, align=left, inner sep=10pt] at (0,1)
        {\fontText
            \textbf{Output (classifier \& xai)} \\[1em]

            \textcolor{customRed}{``step\_by\_step\_thinking": ``To classify whether a patient has diabetes or not, we need to compare the given patient's features with the diagnostic criteria for diabetes. The documents provided do not give specific criteria for diagnosing diabetes, but they do provide information about what to consider when diagnosing diabetes. The relevant documents for this task are Document [2] (Title: Prediabetes -- Introduction), Document [5] (Title: Diabetes Mellitus Screening -- Definition/Introduction), Document [6] (Title: Diabetes Mellitus Screening -- Definition/Introduction -- Type 2 Diabetes), and Document [8] (Title: Mean -- Definition/Introduction). Based on the given patient's features, we can compare them to the diagnostic criteria for diabetes and prediabetes. The patient's age, pregnancies, blood pressure, skin thickness, insulin, BMI, and diabetes pedigree function are all relevant features to consider. The normal blood glucose level is between 70 mg/dL to 99 mg/dL. In patients with prediabetes, blood glucose levels are elevated between 110 mg/dL to 125 mg/dL. The patient's glucose level is 187.0, which is higher than the diagnostic criteria for prediabetes. Therefore, based on this single feature, we can classify the patient as having diabetes."} \\ [1em]
            
            \textcolor{customBlue}{``classification": 1} \\[1em]
            
            \textcolor{customGreen}{``feature\_importance\_ranking": [``Glucose"]} \\[1em]

            \textcolor{customViolet}{``rules\_applied": \{``Glucose": ``Glucose $>$ 125 mg/dL (Diabetes)"\}} \\
             };

\end{tikzpicture}
    
    }
\caption{An example instance from the PIMA dataset is presented. In the table on the top left, we display the patient features along with the classifier's prediction $\tilde{y_i}$ and the enhanced prediction $y_i^r$ generated by the RAG model. The top right section shows the feature importance values calculated using LIME, while the bottom section illustrates the output of the Enhancer module based on the RAG model.}
\label{fig:explanation}
\end{figure} 

\end{document}